\documentclass[a4paper]{article}
\usepackage{amssymb}
\usepackage{graphicx,epsfig}
\usepackage{listings}
\usepackage{rotating}
\usepackage{subfigure}
\usepackage[affil-it]{authblk}

\usepackage{color}
\usepackage{alltt}
\usepackage{verbatim}
\usepackage{url}
\usepackage[latin1]{inputenc}

\usepackage{url}
\urldef{\mailsa}\path|rhgarcia@fidesol.org|
\urldef{\mailsb}\path|pgarcia@atc.ugr.es|
\urldef{\mailsc}\path|jmerelo@geneura.ugr.es|

\begin{document}

\title{Emerging archetypes in massive artificial societies for literary purposes using genetic algorithms}

\author{R.H. Garc\'ia-Ortega
  \thanks{Email: \mailsa, corresponding author}}
\affil{Fundaci\'on I+D del Software Libre, Granada, Spain}
\author{P. Garc\'ia-S\'anchez and J.J. Merelo
  \thanks{Email: \mailsb, \mailsc}}
\affil{GeNeura Team, http://geneura.org, Dept. of Computer Architecture and Technology, University of Granada, Spain}

\maketitle


\begin{abstract}


The creation of fictional stories is a very complex task that usually
implies a creative process where the author has to combine characters,
conflicts and plots to create an engaging narrative. This work
presents a simulated environment with hundreds of characters that
allows the study of coherent and interesting literary archetypes (or
behaviours), plots and sub-plots. We will use this environment to
perform a study about the number of profiles (parameters that define
the personality of a character) needed to create two emergent scenes
of archetypes: ``natality control'' and ``revenge''. A Genetic Algorithm (GA)
will be used to find the fittest number of profiles and parameter
configuration that enables the existence of the desired archetypes
(played by the characters without their explicit knowledge). The
results show that parametrizing this complex system is possible and
that these kind of archetypes can emerge in the given environment.

\end{abstract}


\section{Introduction}
\noindent

In videogames, Non Player Characters (NPCs)  are a type of characters
that live in the game world to provide a more inmersive
experience and, in some cases, present a challenge to the human player. Modern RPGs (Role Playing Games), such as The
Witcher\texttrademark~or Skyrim\texttrademark~ include hundreds of NPC characters. The effort to create a good interactive fiction script is directly proportional to the number of these characters. That is the reason this kind of agents usually counts with limited behaviours, such as wandering in the villages, selling groceries or guarding the cities. Also, they usually offer scripted conversations, for example, to buy and sell objects to the player. In other cases they interact with the player depending of the player's behaviour: for example, if the player steals something a city guard would attack him.  However, these characters do not interact among them, only with the player, and their activities are only guided with this purpose. In a world with such a number of characters, their collective interactions could improve the gaming experience, leading to a richer and more inmersive world. For example, hungry inhabitants could become thieves, guards could pursuit the thieves, villagers could fell in love with others or different war alliances could emerge.

These facts have motivated us to develop a multi-agent system called
MADE (Massive Artificial Drama Engine) to model a self-organized
virtual world where their elements influences each other, following a
cause-effect behaviours in a coherent manner. This system needs to be
a suitable environment for the plot of a specific literary work, being
also interesting for the player/spectator. A set of probabilities and
states are associated to agents' actions, and these probabilities are
optimized by means of an Evolutionary Algorithm (EA) to match with a
specific literary archetype, defined by the fiction creator. The {\em
archetypes} are behaviours and patterns universally accepted and
present in the collective imaginary \cite{ArchetypesGarry05}, that
allows empathize with the characters and immerse yourself in the story
(for example, the well-known {\em hero} archetype).

In this work, several experiments have been carried out to answer the following questions:

\begin{itemize}
 \item Is it possible to model a virtual environment inhabited by hundred of characters with interesting auto-generated behaviour based on literary archetypes?
 \item Could the personality of the agents be parametrized to obtain different behaviours?
 \item How many profiles (groups of parameters that define a personality) are necessary to generate emergent quality sub-plots?
 \item Could a Genetic Algorithm be used to find the fittest parameter values that allow the creation of this kind of sub-plots?
\end{itemize}


In this paper we prove that EAs, together with a proper design of literary patterns, can be used to find the parameters that promote the generation of drama plots and sub-plots in a multi agent based environment.\\

The rest of the work is structured as follows: after the state of the art, the developed system is presented in Section \ref{sec:made}. Then, the experiments conduced with the EA are shown (Sections \ref{sec:experimentalsetup} and \ref{sec:results}). Finally, conclusions and future works are discussed.


\section{State of the art}
\label{sec:soa}


Auto-generated interactive fiction research is mainly focused in methods to create the process of a story generation \cite{nairat2011character}. Story generation can be divided in two areas: interactive and non-interactive. In the first area, and according to \cite{ReviewArinbjarnar09}, an Interactive Drama is defined in a virtual world where the user has freedom to interact with the NPCs and objects in a dramatically interesting experience, different in each execution, and adapted to the interactions of the user.

The generation of interactive dramas can also be based in script
structure \cite{ArchitectureYoung04}, where each possibility in the
story must be previously defined, so there is a limited number of
possible plot combinations.

On the other side, in non-interactive plot generation systems the user does not take control as the protagonist. For example, in the system presented by Pizzi et al. \cite{pizzi2007interactive} the user can interact with the characters, changing their emotions, but making the user an spectator, rather than an actor.

As opposed to those concepts, MADE is focused in Artificial Non Interactive Drama, because its aim is the massive generation of plots for secondary characters, to provide a context for the writer and the player to perceive a virtual world as coherent, detailed and enriched. The story generation (that is, the narrative) is not addressed by MADE, but it has been studied in the systems presents in the survey by Arinbjarnar et al. in \cite{ReviewArinbjarnar09}.


Previous works define the plot as an emergence for the behaviour of the agents that follow a set of rules. In MADE, the agents' behaviour is product of its personality and the environment. That is, the agents does not follow the plot, but they generate the plot itself.

Furthermore, the previous works generate plots in worlds with a limited number of characters. This restriction does not exist in MADE, where the number of characters to create is unlimited.

Following the ideas of the work of Epstein and Axtell
\cite{epstein1996growing} an environment based in {\em Sugarscape} has
been developed with concepts such as food, metabolism and
vision. This environment uses the elements by Gershenson
\cite{gershenson2005general}: a virtual world, agents who are born, grow,

interact, reproduce and dead; resources (food), mediators, and
relations of rivalry (friction) and cooperation (synergy). The actions
of these agents are parametrized according the work of Nairat
\cite{nairat2011character}, based in the use of genetic algorithms to
obtain a plot (solution) where two characters interact in a creative way.

According to the taxonomy described by Togelius et al. \cite{Togelius2011}, the present work
with MADE can be seen has a \textit{procedural content generator} (\textit{PGC})
mainly related to \textit{optional content}, with \textit{stochastic generation}
and modelled as a \textit{generate-and-test} algorithm (search based) that
performs the optimizations of the process during the game development (\textit{offline}).



\section{The MADE Environment}
\label{sec:made}

The MADE environment is a virtual place where different agents play their artificial lives. Its functions are:

\begin{itemize}
\item \textbf{Create an initial set of agents:} MADE environment
  initializes a set of just born orphan agents, each with a profile
  sequentially assigned.
 These agents must compete or collaborate in order to survive.
\item \textbf{Place agents in the map:} the environment has a squared map, formed by cells that can be occupied by one (and only one) agent. The environment allows the agents to discover and interact with other agents in the neighbourhood.
\item \textbf{Start and control the time:} after the creation of the initial set of agents, the MADE environment starts the timer, day by day until a maximum date is reached.
\item \textbf{Execute each agent during a time unit (a day):} In each iteration the list of agents is randomly reordered, and after that following the new order, each agent perform an iteration of its life-cycle, and the dead agents are removed from the grid.
\item \textbf{Perform as an external agent that changes th environment:} In each iteration in the MADE environment, food rations are placed in random cells. An agent only can eat if it is over a cell with a ration, so agents could move the other forcibly.
\item \textbf{Offer services to the agents:} MADE environment allow the agents to check which closer cells have food, are occupied, who agents are in a near position or which positions can be occupied.
\item \textbf{Decide the profile of the agents:} MADE allows the existence of different agent profiles, as previously said. A {\em profile} is a set of characteristics which governs the agent's behaviour.
\end{itemize}

The MADE environment can be configured by using the following parameters (and its values by default): Number of agents initially placed (15), map square grid dimension (10), number of rations randomly placed in the grid each day (10) and duration in virtual days of the  execution of the environment (1000). Those parameters can affect directly to the behaviour of the agents.



\subsection{MADE Agent}
A MADE Agent lives in a MADE Environment, occupies a cell in the grid, moves around looking for food or mate and interacts with other agents.\\

The design of a MADE Agent has some restrictions:
\begin{itemize}
\item An array of parameters (probabilities represented as real numbers between 0 and 1) should be provided in its initialization. These parameters should be used by the agent in its day-by-day decisions.
\item Every action or event is subject to be logged for later analysis.
\item As a result of some iterations of the agent's day-based life cycle, the agent should present the behaviour of a \textit{living thing}, that is born, eats, grows, reacts, reproduces and dies.  
\end{itemize}

\begin{figure}
\begin{center}
\includegraphics[scale=0.65]{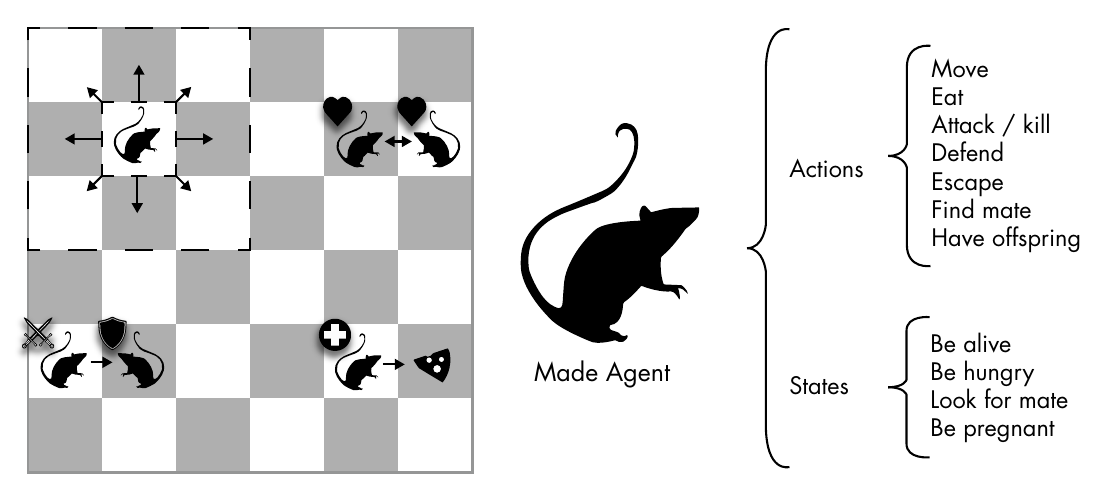}
\caption{Actions and states modelled in the MADE Agent.}
\label{fig:madeAgent}
\end{center}
\end{figure}

A very simple agent has been designed for this study: a virtual
rat, that models:
\begin{itemize}
\item 4 states (be alive, be hungry, look for
mate and be pregnant) that represent internal situations that will lead the
agent to perform the actions described in the item bellow.
\item 7 actions (move, eat, attack, defend, escape,
find mate and have offspring) that lead to a basic instinctive animal
behaviour, very useful for this work since it can be the canvas of 
complex \textit{humanized} behaviour patterns.
\item parameters that define its characteristics and probabilities to
perform actions depending on the state.
\end{itemize}
It's important to remark that no ``feelings'' and no ``memory''
have been modelled in the MADE agent for this study. This is
illustrated in Figure~\ref{fig:madeAgent}.

\begin{figure}
\begin{center}
\includegraphics[scale=0.65]{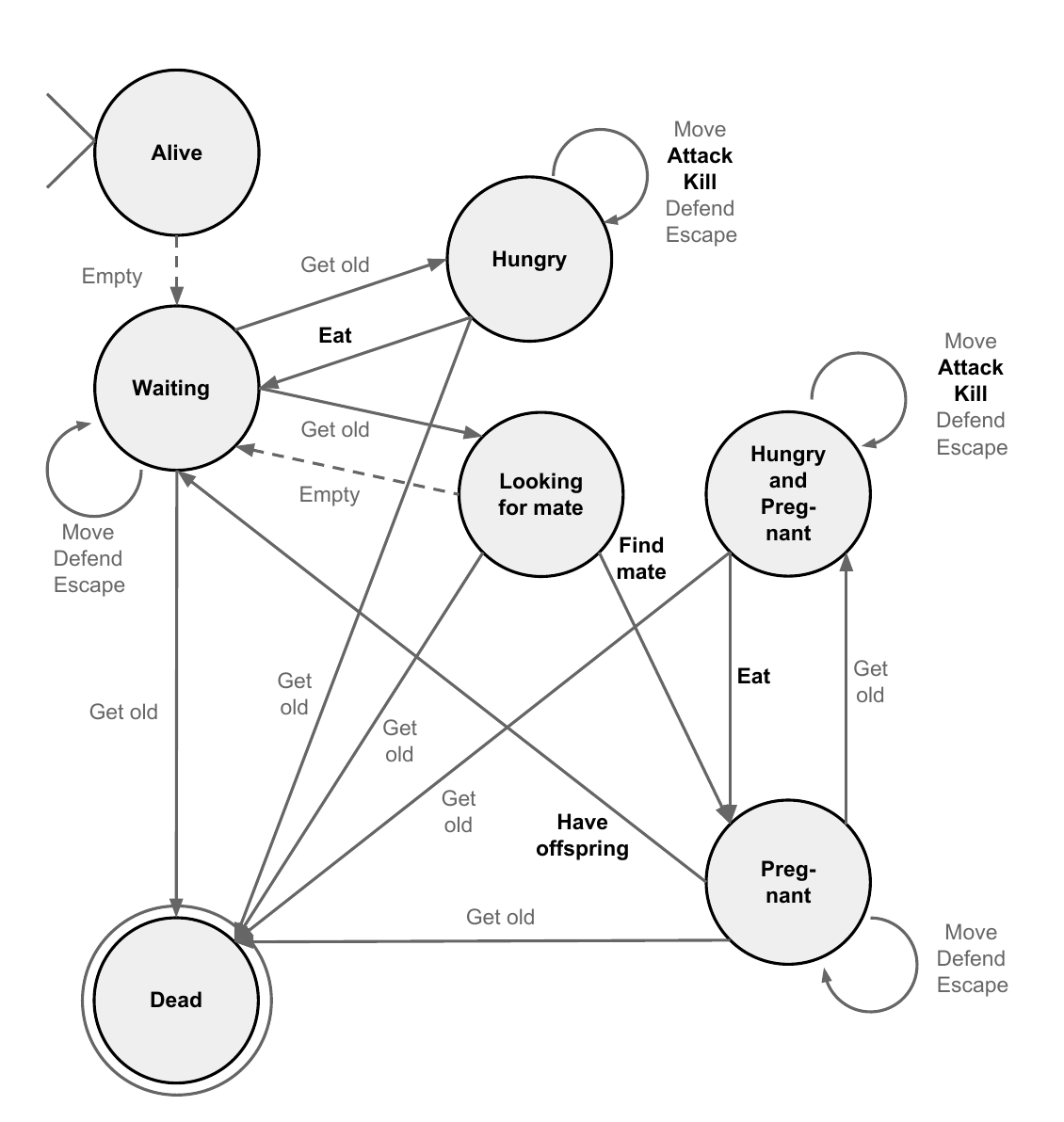}
\caption{MADE Agent's finite state machine}
\label{fig:fsm}
\end{center}
\end{figure}

Every
decision made by the agent is based on its state and its
characteristics (probabilities to perform different actions). 




The MADE Agent is created using twelve parameters, that define its
base features and probabilities to make the decisions presented in the
state diagram in Figure \ref{fig:fsm}. The execution of an
agent is dynamic, and depends on the internal probabilities and states
but also on the neighbourhood, and the map configuration. Even so, we
can say that these initial parameters define in some way the possible
situations where the agent could be involved.

\subsection{MADE Execution}

Every agent has a log that stores all the relevant events in its life in a simple format. Each line of the log indicates the day, the event in a short readable format and some extra information. Every agent's log in the MADE Environment is coherent with all others' logs. Many plots are being created, with a structured format that can be read by game engines or natural language processors, but, given the simplicity of the modelled agents, the stories could be seen by the reader as non-interesting. This log can be used for evaluation. In the next section we propose a method based on EA to let the author of a story promote different behaviours in the MADE agents that could be seen as literary archetypes (usually associated with human feelings and high level cognitive and memory abilities) that can be used to model NPCs in videogames (or be used in other creative areas).


\section{Experimental setup}
\label{sec:experimentalsetup}

Thanks to the agents' logs, we can know every event (internal and
external) of their lives, and evaluate their interest or
adequacy to a specific literary setting.
In this work, we have implemented a method based in regular expressions with backreferences. The proposed technique puts annotations in every agent whose log matches a complex regular expression able to find emerging high level behaviours, not implemented in the life-cycle.

In this proposal, the parameters used to define an agent, mentioned in Section~\ref{sec:made}, are mapped into a chromosome, and a Genetic Algorithm is used to evolve the solution. The fitness function is expressed in terms of:

\begin{itemize}
\item \textbf{Regular expressions applied to the log of each agent in the environment:} An agent is tagged when a regular expression matches its log.
\item \textbf{A numeric function over the number of tagged agents for each archetype:} the fitness of the solution is incremented with the returning value.
\end{itemize}

Different number of profiles (from 1 to 5) have been used to assign
different parameters to different agents. If only one profile is used
in a run, all the agents are created with the same parameters, evolved
by the Genetic Algorithm. If more profiles are used, they are assigned
to the agents in order of appearance in a loop. Our assumption is that
some archetypes could emerge using one profile and other will need
more (those that require two clearly differentiated roles). It's important to remark that the number of alleles of the chromosome are multiplied by the number of profiles, so the convergence of the solution could be affected by the number of profiles used.

For the experiments performed in this work, we have used the parameters shown in Table~\ref{fig:ga_parameters}. These values have been chosen empirically after several test runs.

\begin{table}
\begin{center}
\caption{Parametrization of the Genetic Algorithm}
\label{fig:ga_parameters}
\begin{tabular}{p{3cm}p{7cm}}
\hline\noalign{\smallskip}
\noalign{\smallskip}
Parameter & Value \\
\hline
\noalign{\smallskip}
Codification & 12 alleles per profile\\
Fitness function & Average of 10 executions.\\
Natural selector & Original Rate: 0.9 \\
Crossover operator & Rate: 35\% \\
Mutation operator & Desired Rate: 12 \\
Stop condition & 100 executions\\
Generations & 30\\
Population size & 30 \\
\hline
\end{tabular}

\end{center}
\end{table}


For this work, two sample groups of archetypes (or scenes)  have been
chosen: 
The first one is called ``natality control'' and its goal is to model
a global archetype (population growth) and four individual memory
based archetypes (\textit{downtrodden}, \textit{helpless},
\textit{warrior} and \textit{bad warrior}; these are related to their
attack and defense behaviors).
 The second scene is called
``revenge'' and its goal is to model an individual complex memory
based behaviour between two characters (\textit{revenge}).

The Experiment 1, ``Natality control'' scene, aggregates different sample archetypes where many factors must be taken into account.  It tries to find what number of profiles and values are optimal to:
\begin{itemize}
\item Ensure that, after 1000 virtual days, the alive population will be the 60\% of the total population. This is called a \textit{global archetype}.
\item Emerge the \textit{downtrodden} archetype in the 22\% of the
  population. An agent will be considered as a \textit{downtrodden} or
  \textit{defender} if it has been attacked at least two times and has
  defended the position.

\item Emerge the \textit{warrior} archetype the 22\% of the population. An agent will be considered as a \textit{warrior} if it has satisfactory attacked at least five times.
\item Emerge the \textit{helpless} archetype the 22\% of the population. An agent will be considered as a \textit{helpless}  if it has been attacked at least ten times and hasn't defended the position.
\item Emerge the \textit{bad warrior} archetype the 22\% of the population. An agent will be considered as a \textit{bad warrior}  if it has unsatisfactory attacked at least ten times.
\end{itemize}
We have used the presented values to define this scene because, as our opinion, they model an interesting literary scene.

To model this scenes we have defined the fitness function as follows:
If the exact percentage of agents are tagged with one archetype defined, 1 point is added to the fitness. The maximum is therefore, 5 points. However, all the fitnesses use a normal distribution over the percentage of appearance. For example, for the first scene of archetypes (``growing population''), the maximum value (1) is obtained when the 60\% percent of the population is alive, and the normal distribution begins in the 30\% and ends in the 90\%. For the rest of the archetypes, 1 point is added if the 22'5\% of the population is tagged with each one of the archetypes, and each normal distribution begins in the 8\% and ends in the 30\%.


The experiment 2, ``Revenge archetype'' scene, is performed to make more complex memory based behaviour emerge between two characters:  It tries to find what number of profiles and values are optimal to make \textit{revenge} archetype emerge in as many agents as possible after 1000 days.  An agent (a) will be considered as a \textit{avenger} if it has been attacked by other agent (b) and after that, in a moment in its life, it has satisfactory attacked the agent b, in revenge. The value of the days is set to 1000 because is a duration long enough to make the archetype emerge.

We have defined the fitness function of this experiment as follows:
For each agent, if the agent's log matches the archetype, it adds 1 point to the fitness. Therefore, the goal is to recreate an environment where most ``avenger'' agents exist.

The available source code of the MADE environment and the algorithms used in this experiment are publicly available in \url{https://github.com/raiben/made} under a LGPL license. 

\section{Results and discussion}
\label{sec:results}

\begin{figure}[htb]
\centering

\subfigure[Experiment 1]{
   \includegraphics[width=13.5pc] {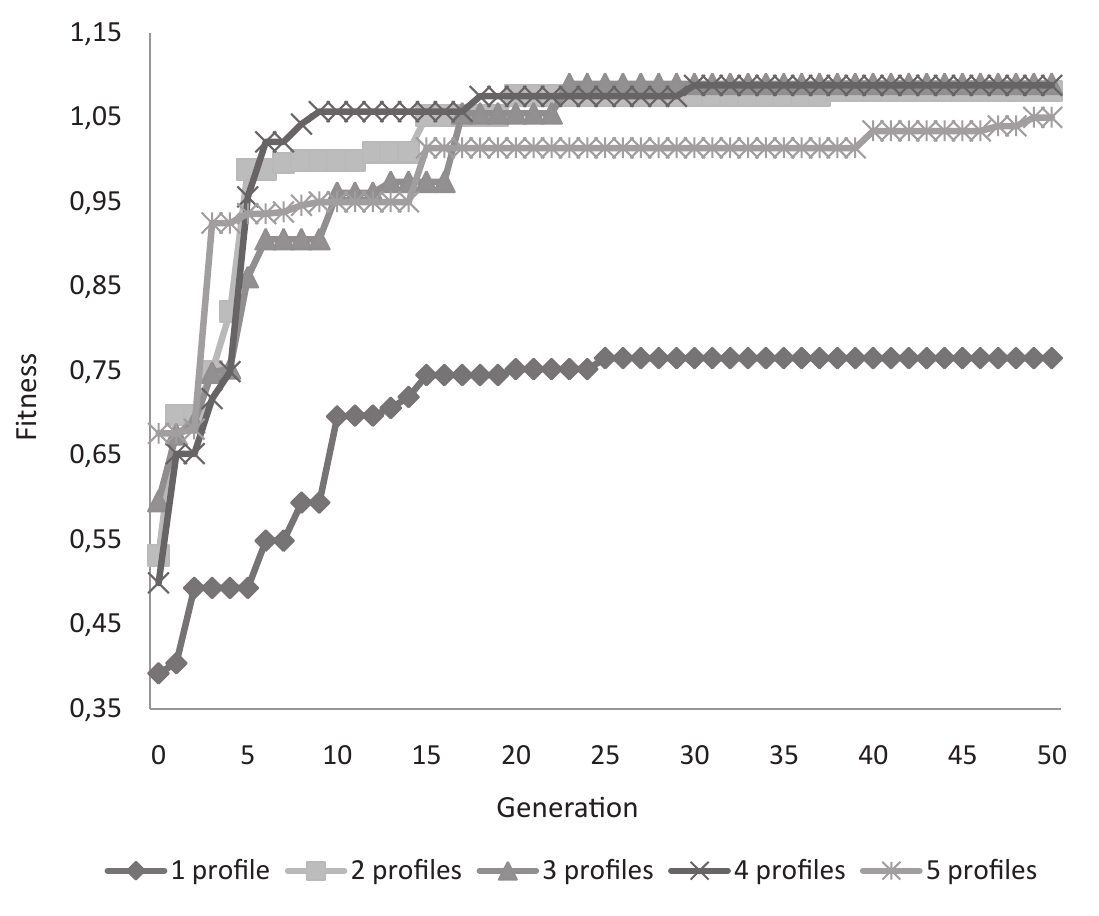}
   \label{fig:evo1}
 }
\subfigure[Experiment 2]{
   \includegraphics[width=13.5pc] {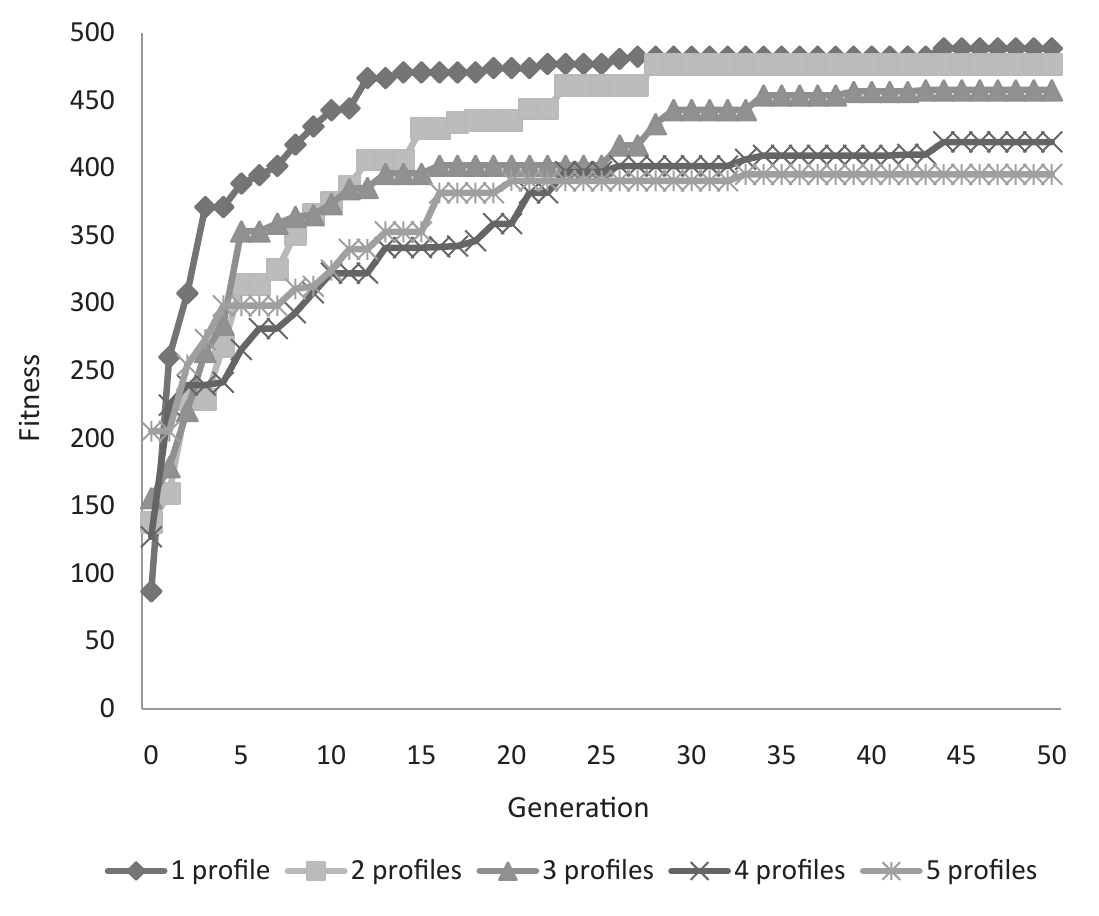}
   \label{fig:evo2}
 }
\caption{Example of the evolution of the best individual for the five
  profiles for each configuration.}

\label{fig:boxplots}
\end{figure}

Table~\ref{fig:exp1_30ex} shows the average of the best fitness and the average population fitness at the end of each execution for each configuration: number of profiles from 1 (P1) to 5 (P5) in the experiment 1 (``natality control'').
The evolution of the best fitness for each configuration is shown in Figure~\ref{fig:evo1}. We have performed a Kruskal-Wallis test for the best individuals fitness, obtaining differences among all the number of profiles (p-value $<<$0.05). As we suspected, it is clear that using one profile is not enough to emerge the desired archetype. However, the pairwise comparison using Wilcoxon does not find significant differences using more than 2 profiles. This can be explained because an agent could share more than one archetype at the same time.  A promising number of profiles could be 4, because their only lower outlier is not as distributed as the others. As can be seen in Figure \ref{fig:evo1} the evolution of the best fitness increases with all possible number of profiles, existing therefore an increase of the performance of the system.

\begin{table}
\begin{center}
\caption{Results for 30 executions of each configuration using 1 to 5 profiles in the experiment 1}
\label{fig:exp1_30ex}

\begin{tabular}{lllll}
\hline\noalign{\smallskip}
\parbox[t]{2cm}{Number of\\ profiles}
& \parbox[t]{3cm}{Best fitness\\ (average) *}
& \parbox[t]{3cm}{Average\\ fitness **} \\
\noalign{\smallskip}
\hline
\noalign{\smallskip}
1 & 0,765 $\pm$ 0,037 & 0,761 $\pm$ 0,038 \\
2 & 1,063 $\pm$ 0,115 & 1,059 $\pm$ 0,114 \\
3 & 1,093 $\pm$ 0,063 & 1,091 $\pm$ 0,062 \\
4 & 1,084 $\pm$ 0,048 & 1,082 $\pm$ 0,048 \\
5 & 1,045 $\pm$ 0,110 & 1,041 $\pm$ 0,108 \\
\hline
\end{tabular}
\\
\** Average of the best fitness at the end of each execution\\
\*** Average population fitness  at the end of each execution \\
\end{center}
\end{table}

\begin{figure}[htb]
\centering

\subfigure[Experiment 1]{
   \includegraphics[width=13.5pc] {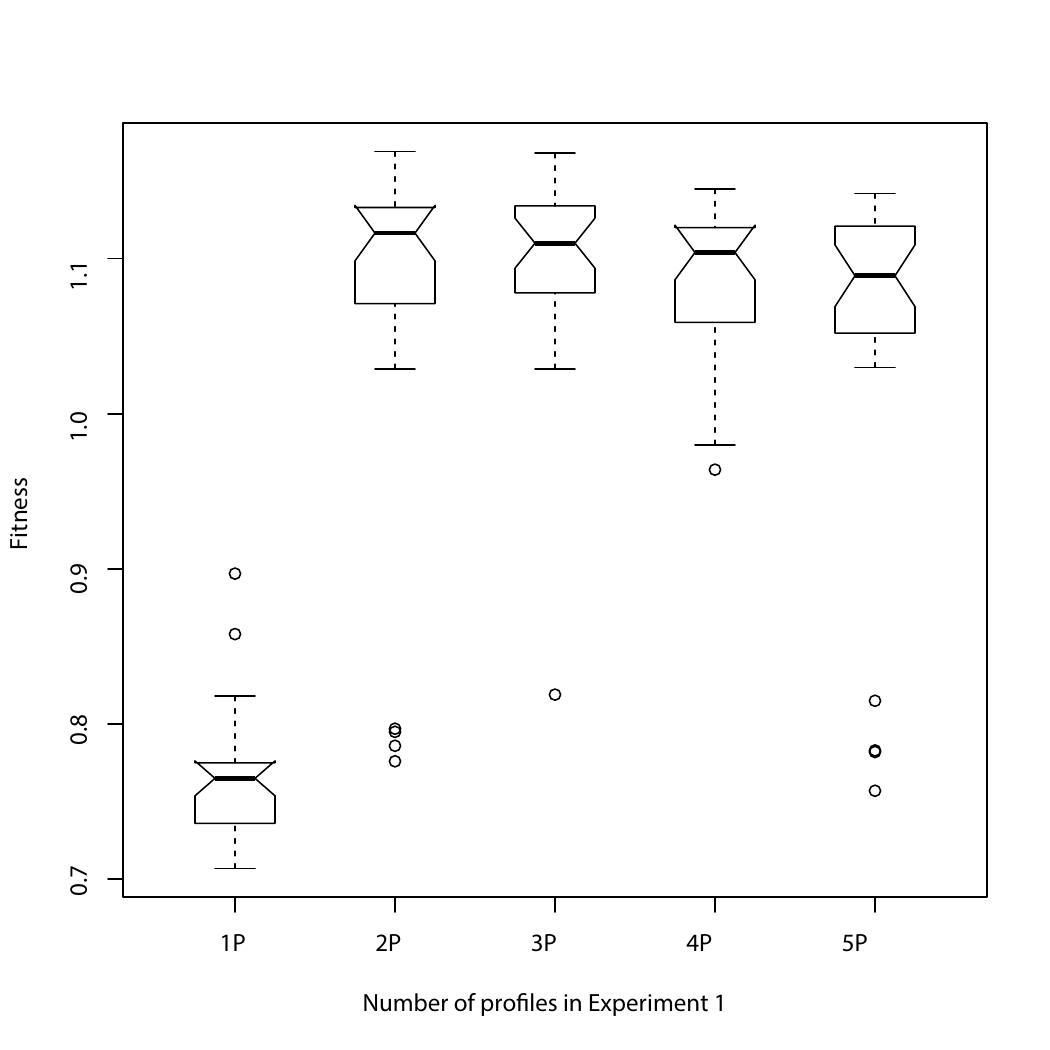}
   \label{fig:subfig1}
 }
\subfigure[Experiment 2]{
   \includegraphics[width=13.5pc] {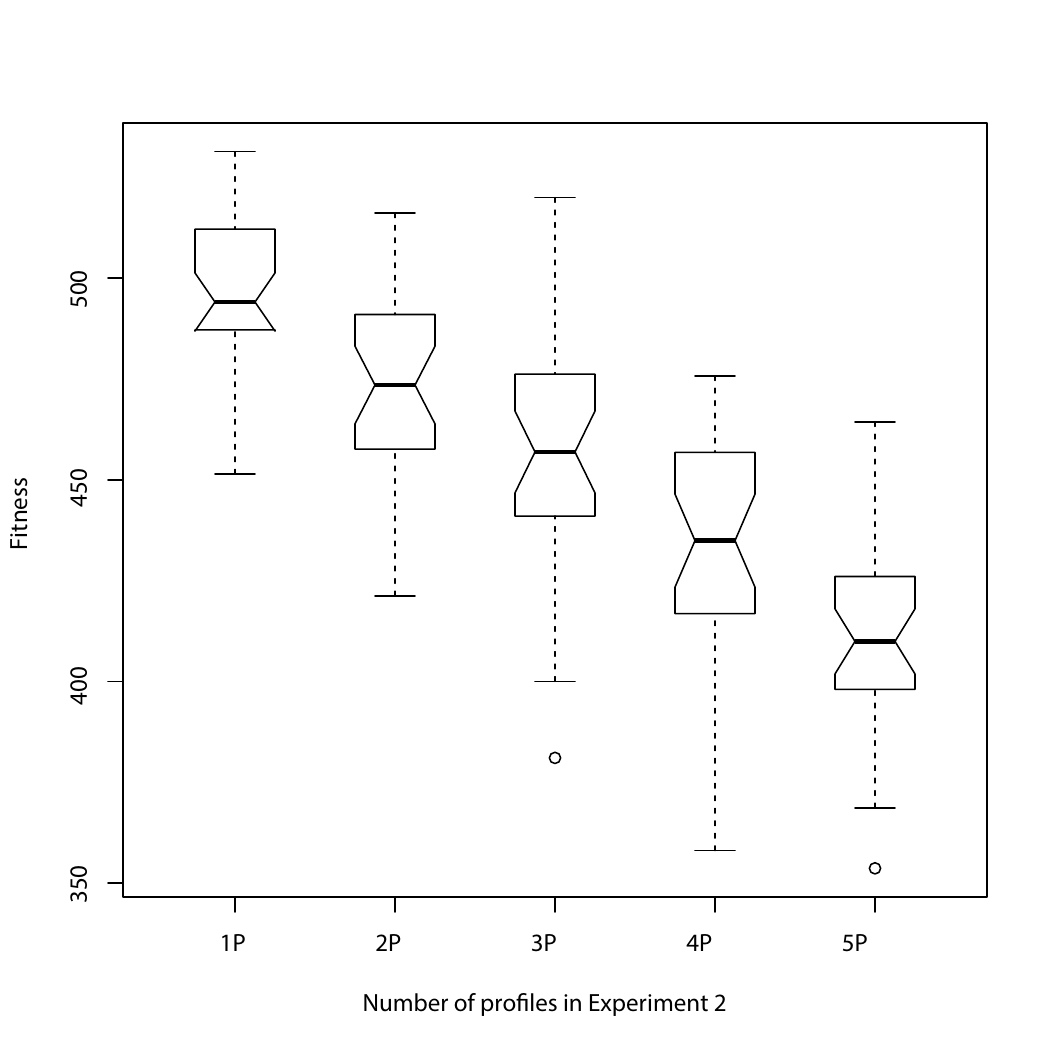}
   \label{fig:subfig2}
 }
\caption{Average fitness of the 30 best individuals for each configuration.}

\label{fig:graph}
\end{figure}

The results of the second experiment (``revenge'') are shown in Table~\ref{fig:exp2_30ex}. It shows the average of the best fitness and the average population fitness at the end of each execution for each configurations. Boxplots of the best fitness obtained are shown in Figure \ref{fig:subfig2}. In this case, Krukal-Wallis and Wilcoxon pairwise comparison shows significant differences among all configuration (p-value $<<$ 0.05) except between P2 and P3 (p-value=0.3). Therefore, we can conclude that in this kind of global archetype only a profile must be used for obtaining the best results. This makes sense, because we are looking for one type of local archetypes ({\em avenger}), so adding extra profiles leads to different behaviours of the agents.

\begin{table}
\begin{center}
\caption{Results for 30 executions of each configuration using 1 to 5 profiles in the experiment 2}
\label{fig:exp2_30ex}
\begin{tabular}{lllll}
\hline\noalign{\smallskip}
\parbox[t]{2cm}{Number of\\ profiles}
& \parbox[t]{3cm}{Best fitness\\(average) *}
& \parbox[t]{3cm}{Average\\fitness **} \\
\noalign{\smallskip}
\hline
\noalign{\smallskip}
1 & 495,513 $\pm$ 20,091 & 493,908 $\pm$ 19,884 \\
2 & 471,206 $\pm$ 24,550 & 469,361 $\pm$ 24,015 \\
3 & 455,42 $\pm$ 28,240 & 452,787 $\pm$ 29,438 \\
4 & 431,926 $\pm$ 31,682 & 428,206 $\pm$ 31,238 \\
5 & 411,24 $\pm$ 25,023 & 408,387 $\pm$ 23,829 \\
\hline
\end{tabular}
\\
\** Average of the best fitness at the end of each execution\\
\*** Average population fitness  at the end of each execution \\
\end{center}
\end{table}

\section{Conclusions}
\label{sec:conclusion}

This work presents the result obtained by the MADE environment,
 a self-organized multi-agent
system to model complex societies which can be used to study
emergent behaviours for literary purposes; for example, to extract
interesting plots for NPCs in videogames. In this paper we have used
the MADE environment to establish the number of profiles (sets of
personality parameters) necessary to emerge two different global
scenes: ``natality control'' and ``revenge'',
which have been chosen because of their different nature: the first
one consist in specific archetype rates (related to defense and attack
behaviours of the agents) inside a global population rate, while the second one
consist in maximizing the appearance of a memory based archetype, like the
revenge (as previously said, memory hasn't been explicitly modelled in the
agent).

We have used a
Genetic Algorithm to optimize the parameters of these profiles using
as a fitness a function that model the two desired global
archetypes. Results show that in the first archetype the number of
profiles must be at least two, because it is necessary that different
behaviours (local archetypes) emerge. On the contrary, on the second
experiment, using only one profile is the best configuration, because
the purpose was to emerge only one archetype ({\em avenger}). If the
number of profiles is increased worse results are obtained.

This implies that each scene or combination of desired archetypes for a given
environment can be mapped to a number of profiles that maximize the appearance
of these archetypes, and that the fittest number of profiles and its values can
be obtained by using Genetic Algorithms.  
Given a literary setting, an author of a story or a videogame could define 
different rates of archetypes or behaviour patterns and use the techniques described in
the present work to obtain the optimal profiles. The execution of the MADE Environment
using these profiles as input would produce a background (or set of characters' lives)
where the archetypes have emerged and have automatically created massive plots coherent
with the settings of the artwork.

In future works, more complex agents will be used, with different
rules to be modelled. For example, we plan to model more human
behaviours such as love or envy, to generate interesting plots such as
wars, weddings, or family crimes. Different fitnesses will be used,
for example, taking into account human opinions to establish the
interestingness of a generated plot. Also, this system will be tested
into an existent and well-known game, such as Skyrim\texttrademark,
whose AI engine is publicly available for players and researchers. 


\section*{Acknowledgements}
This work has been supported in part by FPU research grant AP2009-2942 and projects EvOrq (TIC-3903), CANUBE (CEI2013-P-14) and ANYSELF (TIN2011-28627-C04-02).

\bibliographystyle{splncs}
\bibliography{made}

\begin{thebibliography}{1}

\bibitem{ArchetypesGarry05}
Garry, J., El-Shamy, H.M.:
\newblock Archetypes and Motifs in Folklore and Literature: A Handbook.
\newblock M.E. Sharpe (2005)

\bibitem{nairat2011character}
Nairat, M., Dahlstedt, P., Nordahl, M.G.:
\newblock Character evolution approach to generative storytelling.
\newblock In: Evolutionary Computation (CEC), 2011 IEEE Congress on, IEEE
  (2011)  1258--1263

\bibitem{ReviewArinbjarnar09}
Arinbjarnar, M., Barber, H., Kudenko, D.:
\newblock A critical review of interactive drama systems.
\newblock In: AISB 2009 Symposium. AI \& Games, Edinburgh, Citeseer (2009)

\bibitem{ArchitectureYoung04}
Young, R.M., Riedl, M.O., Branly, M., Jhala, A., Martin, R., Saretto, C.:
\newblock An architecture for integrating plan-based behavior generation with
  interactive game environments.
\newblock Journal of Game Development \textbf{1}(1) (2004)  51--70

\bibitem{pizzi2007interactive}
Pizzi, D., Charles, F., Lugrin, J.L., Cavazza, M.:
\newblock Interactive storytelling with literary feelings.
\newblock In: Affective Computing and Intelligent Interaction.
\newblock Springer (2007)  630--641

\bibitem{epstein1996growing}
Epstein, J.M., Axtell, R.L.:
\newblock Growing Artificial Societies: Social Science from the Bottom Up
  (Complex Adaptive Systems).
\newblock The MIT Press (1996)

\bibitem{gershenson2005general}
Gershenson, C.:
\newblock A general methodology for designing self-organizing systems.
\newblock arXiv preprint nlin/0505009 (2005)

\bibitem{Togelius2011}
Togelius, J., Yannakakis, G.N., Stanley, K.O., Browne, C.:
\newblock {Search-Based Procedural Content Generation: A Taxonomy and Survey}.
\newblock IEEE Transactions on Computational Intelligence and AI in Games
  \textbf{3}(3) (2011)  172--186

\end{thebibliography}

\end{document}